\documentclass[journal]{IEEEtran}

\usepackage{graphicx}
\usepackage{amsmath,amsfonts}
\usepackage{float}

\usepackage{array}
\usepackage{multirow}
\usepackage{algorithm, algorithmic}
\usepackage{makecell}
\usepackage{setspace}

\usepackage{amssymb}
\usepackage{booktabs}
\usepackage{caption,setspace}
\usepackage{tabularx}
\usepackage[export]{adjustbox}
\usepackage{tikz}

\usepackage{subcaption}
\usepackage{textcomp}
\usepackage{stfloats}
\usepackage{url}
\usepackage{verbatim}
\usepackage{cite}
\begin{document}

\title{Numerical Weather Forecasting using Convolutional-LSTM with Attention and Context Matcher Mechanisms}

\author{Selim F. Tekin, Arda Fazla and Suleyman S. Kozat,~\IEEEmembership{Senior Member,~IEEE}%
\thanks{This work has been submitted to the Journal of IEEE Transactions on Geoscience and Remote Sensing on July 18th, 2023.}
\thanks{S. S. Tekin is with the Department of Computer Science, Georgia Institute of Technology, Atlanta, USA, e-mail: stekin6@gatech.edu.}
\thanks{A. Fazla, S. S. Kozat are with the Department of Electrical and Electronics Engineering, Bilkent University, Ankara, Turkey, e-mail: arda@ee.bilkent.edu.tr, kozat@ee.bilkent.edu.tr.}
\thanks{The authors would like to thank Turkish State Meteorological Services for providing their extensive experience in weather forecasting. This work is supported by the Turkish State Meteorological Service and in part by the Turkish Academy of Sciences Outstanding Researcher Program.}}

\markboth{Submitted to the Journal of IEEE Transactions on Geoscience and Remote Sensing}
{TEKIN \MakeLowercase{et al.}: Numerical Weather Forecasting using Convolutional-LSTM with Attention and Context Matcher Mechanisms}

\maketitle

\begin{abstract}
Numerical weather forecasting using high-resolution physical models often requires extensive computational resources on supercomputers, which diminishes their wide usage in most real-life applications. As a remedy, applying deep learning methods have revealed innovative solutions within this field. To this end, we introduce a novel deep learning architecture for forecasting high-resolution spatio-temporal weather data. Our approach extends the conventional encoder-decoder structure by integrating Convolutional Long-short Term Memory and Convolutional Neural Networks. In addition, we incorporate attention and context matcher mechanisms into the model architecture. Our Weather Model achieves significant performance improvements compared to baseline deep learning models, including ConvLSTM, TrajGRU and U-Net. Our experimental evaluation involves high-scale, real-world benchmark numerical weather datasets, namely the ERA5 hourly dataset on pressure levels and WeatherBench. Our results demonstrate substantial improvements in identifying spatial and temporal correlations with attention matrices focusing on distinct parts of the input series to model atmospheric circulations. We also compare our model with high-resolution physical models using the benchmark metrics and show that our Weather Model is accurate and easy to interpret.
\end{abstract}


\section{Introduction}

\IEEEPARstart{W}{eather} is a vital scientific research area as it is the cornerstone of every real-life system such as agriculture, energy production, tourism, transport, navigation and climate \cite{JournalRef1, JournalRef2}. Accurate forecasts prevent life and economic losses and support emergency management by diminishing the effects of high-impact weather events and creating substantial financial revenue \cite{intro}. Foundations of modern-day weather forecasting were introduced in the early 20th century when scientists solved a system of nonlinear differential equations by hand \cite{LYNCH20083431}. Today, super-computers solve these equations for millions of points per time step to accurately forecast up to months ahead \cite{300billion}. Meteorologists call this problem numerical weather prediction (NWP), where the objective is to predict a numerical value of a weather feature such as temperature or wind speed for a location. However, NWP models are getting more complex and their demand for high computation power increases continuously \cite{300billion}. Obtaining results from these models can take hours, days or sometimes weeks, limiting their ability to provide actionable predictions in a short time.

We focus on NWP for multiple locations, which is a spatio-temporal time series problem. A spatio-temporal time series contains multiple time series belonging to different locations with temporal and spatial covariances such as the temperature at multiple weather stations in a city. Hence, NWP is a spatio-temporal prediction problem on a grid. In this grid structure, compared to commonly used physical models for NWP, deep learning models can provide results within minutes of receiving data, exploit the big data aggregated in years and make accurate predictions with significantly less computational resources \cite{Rodrigues2018}. The success of spatio-temporal time series forecasting with deep learning has emerged with the introduction of Convolutional Long-short Term Memory (ConvLSTM) cells in \cite{convlstm}, a modified version of Long-short Term Memory (LSTM) cells of \cite{lstm} that is targeted to capture spatial patterns using convolution operations. Specifically, ConvLSTMs focus on the precipitation nowcasting problem, where the task is to give a precise and timely prediction of rainfall intensity in a local region over a relatively short period \cite{ref1, ref2}. Following this model, \cite{convlstm, trajgru} introduce new approaches to model spatial and temporal patterns and show high performance in different tasks such as weather, crime and traffic density forecasting. However, previous studies do not exploit the numerous datasets available for weather such as satellite images, numerical grid values and observations in meteorological stations. Instead, these studies usually focus on a single input source such as a sequence of radar images \cite{JournalRef6, JournalRef7}. The weather, however, is a chaotic system affected by many exogenous factors, e.g., vegetation, geographical contour and human intervention. Therefore, more data sources and novel structures are needed to model such chaotic behaviors \cite{300billion}.

Here, we call our target series the endogenous variable and the different data sources entering the forecasting model as the side-information or exogenous variables. The side-information can be concatenated with the input series and directly given to the model. However, the side-information integration in previous works shows more effective ways \cite{DeVries2017, Perez2018, Wang2018}. \cite{dual_attn} demonstrates that implementing Fully Connected Neural Networks (FC-NNs) before the memory units allows the model to specifically focus on the patterns of the exogenous series that are correlated with the endogenous series. This process is known in the literature as attention and was introduced in \cite{Bahdanau, attention}. Even though FC-NNs are effective, they cannot capture spatial information as well as Convolutional Neural Networks (CNNs). Thus, \cite{Zhang2018} introduced a convolutional attention mechanism for gesture recognition and indicated the need for spatial attention in ConvLSTM structures. Moreover, \cite{Agrawal2019} and \cite{ref2} demonstrated the high performance of the U-net model in weather forecasting, which contains multiple layers of CNNs with skip connections. Similar to \cite{convlstm, trajgru, traj_cv}, U-net follows the idea of encoding input spatio-temporal time series to a small abstract representation and decoding this representation to generate an output sequence. The advantage of an encoder-decoder structure is that the length of the input and output sequences may differ, allowing forecasts of varying length. Recently, \cite{Rasp2020} released a benchmark dataset from the ERA5 archive, introduced evaluation metrics to make inter-comparison between studies and provided baseline scores for multiple baseline deep learning and physical forecasting models. These baseline deep learning models make accurate short-term predictions, however, their performance deteriorates rapidly as the length of the output sequence increases.

\begin{figure*}[!ht]
    \centering
    \begin{subfigure}{0.7\textwidth}
        \centering
        \includegraphics[width=\textwidth]{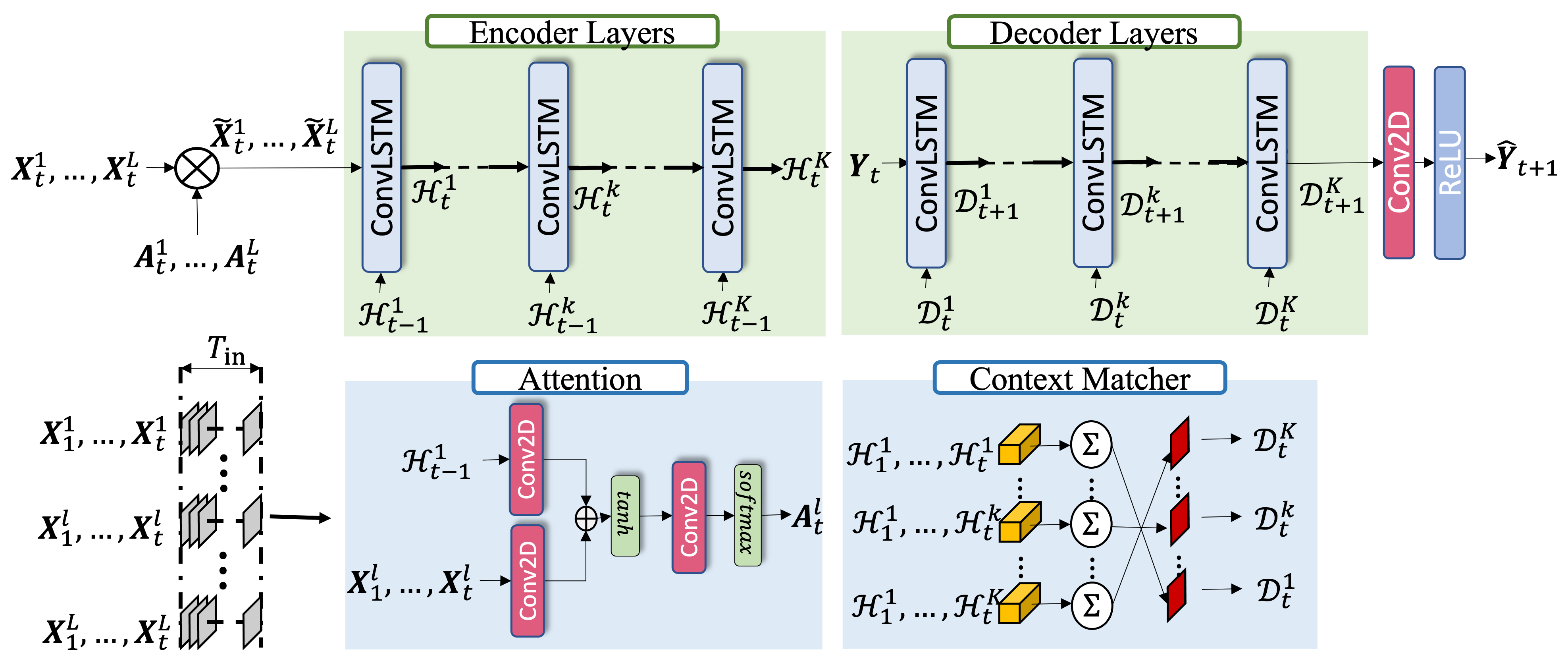}
        \caption{At each time step $t$, we first multiply all weather features $\{\mathbf{X}^1_t, \dots, \mathbf{X}^l_t, \dots,  \mathbf{X}^L_t\}$ with attention matrices $\{\mathbf{{A}}_{t}^{1}, \dots,  \mathbf{{A}}_{t}^{l}, \dots, \mathbf{{A}}_{t}^{L}\}$ to create weighted input series $\{\tilde{\mathbf{X}}_{t}^{1}, \dots, \tilde{\mathbf{X}}_{t}^{l}, \dots, \tilde{\mathbf{X}}_{t}^{L}\}$, where $\tilde{\mathbf{X}}_t^{l}=\mathbf{X}^{t}_{l}\odot \mathbf{{A}}_{t}^{l}$. Our attention mechanism  computes the corresponding attention weight $\mathbf{{A}}_{t}^{l}$, for each given feature. Here, the attention weights are conditioned on the previous hidden state of the encoder's first layer, $\mathcal{H}_{t-1}^1$. Next, the weighted input series enter the first ConvLSTM unit in the encoder. After the input passes all the encoder layers, the encoder's hidden states $\{\mathcal{H}_{1}^{1}, \dots, \mathcal{H}_{1}^k, \dots, \mathcal{H}_{t}^K\}$ enter to the context matcher, where $k$ represents the layer index. The context matcher applies summation to the hidden states and reverses the layer order of the states to create $\{\mathcal{D}^1_{t-1}, \dots, \mathcal{D}^k_{t-1}, \dots, \mathcal{D}^K_{t-1}\}$. Finally, $\mathcal{D}^K_{t}$ passes through output convolutions to create our model prediction $\mathbf{\hat{Y}}_{t+1}$.}
        \label{fig:model_graph}
    \end{subfigure}
    \begin{subfigure}{0.4\textwidth}
        \centering
        \includegraphics[width=\textwidth]{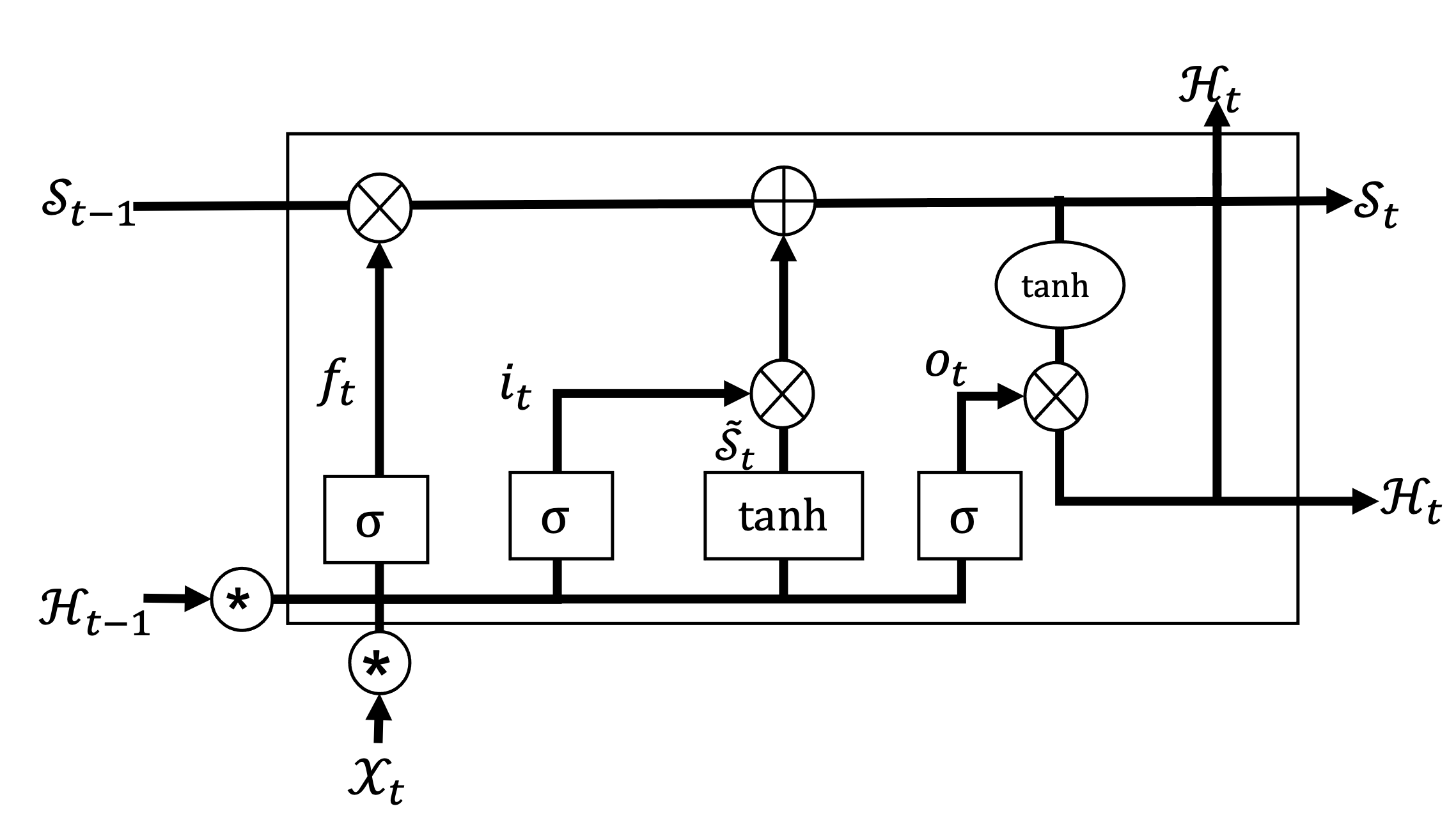}
        \caption{The inside of a ConvLSTM cell, which consists of the input $i_t$, cell $s_t$, forget $f_t$ and output $o_t$ gates along with varying operations in each gate. Joining arrows represent the concatenation of the matrices. Here, we apply convolution operations on $\mathcal{H}_{t-1}$ and $\mathcal{X}_{t}$ and two different hidden states $\mathcal{H}_t$ and $\mathcal{S}_t$ are generated in each time step.}
        \label{fig:convlstm}
    \end{subfigure}
    \begin{subfigure}{0.45\textwidth}
        \centering
        \includegraphics[width=\textwidth]{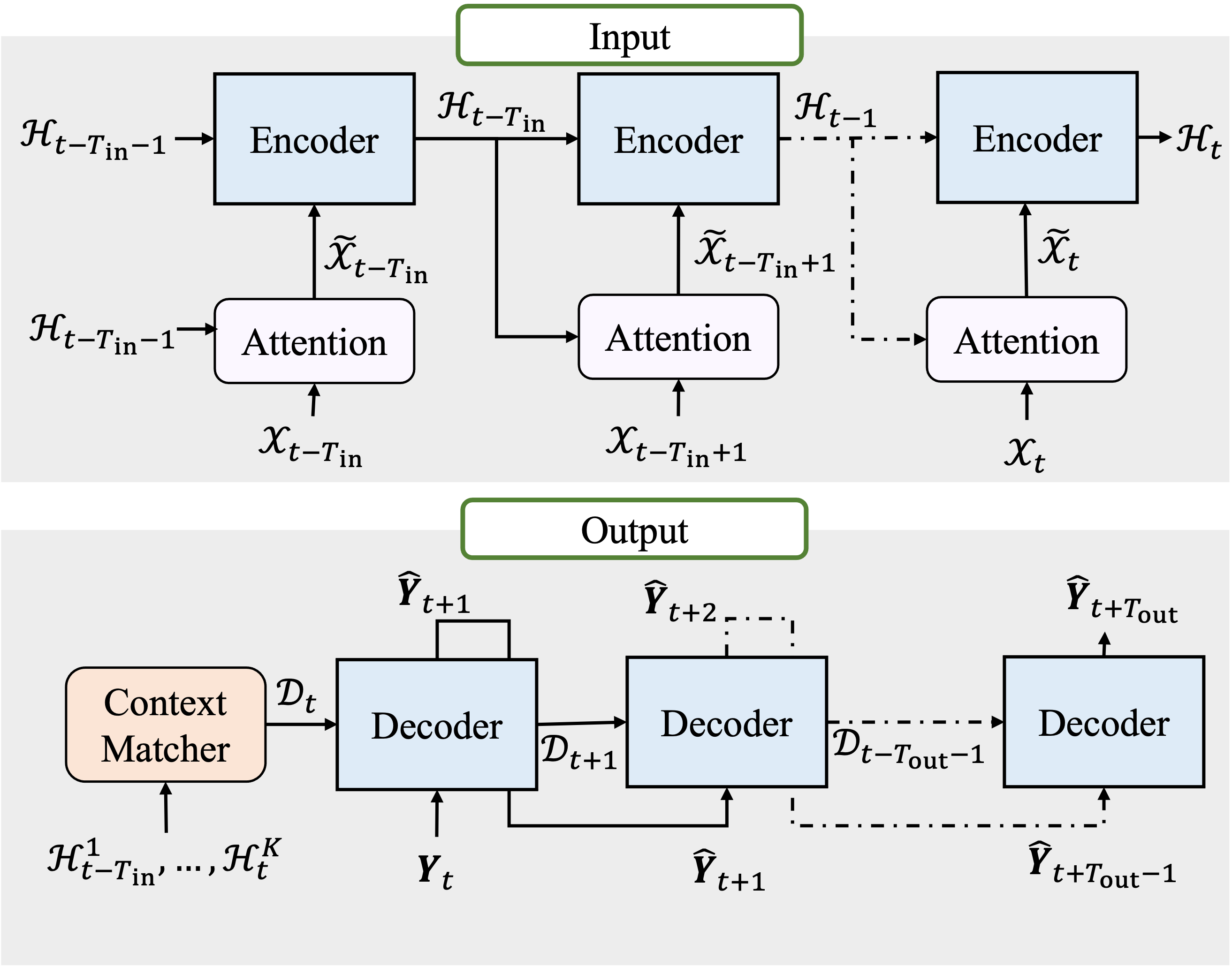}
        \caption{The unrolled view of the encoder and decoder structure. The input sequence first passes to the encoder, which encodes the input sequence to the hidden states. Then, for the first prediction, the decoder takes $\mathbf{Y}_t$ and $\mathcal{D}_t$ to predict $\mathbf{\hat{Y}}_{t+1}$. After that, the decoder uses this prediction and the former state $\mathcal{D}_{t+1}$ to predict the next output $\mathbf{\hat{Y}}_{t+2}$. This recursive process continues until $t=T_{\mathrm{out}}$.}
        \label{fig:enc_dec}
    \end{subfigure}
    \caption{Graph of our Weather Model is demonstrated in order to further clarify the recursive output generation process. (a) The details of the encoder, decoder, attention and context matcher architectures. (b) The inside of a ConvLSTM cell. (c) The unrolled view of the decoder architecture.}
    \label{fig:all_model_fig}
\end{figure*}

To this end, we introduce a novel deep learning architecture for NWP by combining multiple spatio-temporal data sources with an attention mechanism and using ConvLSTM cells as building blocks to capture spatial and temporal correlations. We illustrate our complete model architecture in Fig. \ref{fig:model_graph}. Our model selects relevant cells in multiple spatio-temporal time series to make long-term spatial predictions by preserving the long-term dependencies using attention and context matcher mechanisms. Thus, we introduce an encoder-decoder architecture specific for NWP. First, our attention-based encoder block consisting of stacked ConvLSTM units encodes the input sequence to hidden states. Then, our context matcher mechanism aligns the decoder's hidden states by summing the encoder's hidden states across all time steps. This process addresses long-term dependencies by extending the length of the gradient flows. Each ConvLSTM unit in the decoder block decodes the passed information from the hidden layers of the encoder. The output of the decoder is then passed through the output convolutions to produce predictions. Altogether, our model architecture is adept at focusing on different parts of the spatio-temporal time series and successfully capturing long-term dependencies. We call our architecture the Weather Model (WM). We demonstrate significant performance gains over benchmark deep learning models through extensive experiments on two real-life datasets from the ERA5 archive. Furthermore, we openly share our source code to facilitate further research and comparison.

\subsection{Prior Art and Comparisions}

Applying deep learning methods to weather forecasting is an extensively studied area. \cite{liu2015deep} leveraged a massive volume of weather datasets to train an Auto-encoder with multi-layered neural networks (NN) to forecast hourly weather data. \cite{Hossain2015} improved this approach by introducing stacked auto-encoders to temperature forecasting. Later, \cite{Lee2018} implemented a simple NN to forecast rainfall using historical rainfall data and lagged climate indices from various stations. These works demonstrated the high performance of simple deep learning architectures and their potential with high volume of data. Moreover, \cite{recurrent, lstm_sequence} applied Recurrent Neural Networks (RNNs) and LSTMs to weather forecasting and \cite{Poornima2019} focused on the vanishing gradient problem by implementing an intensified LSTM architecture for rainfall forecasting. \cite{Wang2019} introduced deep uncertainty quantification using an RNN-based architecture and obtained a single-value forecast. Recently, \cite{Hewage2020} used Temporal Convolutional Network (TCN) for multiple-input single-output weather predictions and demonstrated superior results with respect to a physical NWP model, namely the Weather Research and Forecasting model (WRF) \cite{wrf}. However, this approach is costly, does not comprise multiple features and spatial covariances and only provides high performance on multiple-input single-output models.

As a remedy, \cite{convlstm} introduced ConvLSTM and Convolutional Gated Recurrent Units (ConvGRU) to the forecasting network, which is an encoder-decoder structure consisting of stacked ConvLSTMs or ConvGRUs. These models capture the spatial-temporal covariances and demonstrate high performance in rainfall intensity prediction by using radar echo sequences. The authors also showed that ConvLSTM is more accurate than the optical flow model \cite{woo2017operational} and better at capturing spatio-temporal correlations than the fully-connected LSTM. Later, \cite{trajgru} introduced Trajectory Gated Recurrent Unit (Traj-GRU) to learn location invariant movements between input radar images, which achieved slightly higher performance compared to ConvGRU. Although ConvLSTM and TrajGRU offer high performance, they are limited by the encoder-decoder structure, which restricts the output production to the dimensionality of the given input. As a remedy, \cite{Tran2019} improved the encoder-decoder architecture by integrating downsampling and upsampling layers between TrajGRU units and eliminated the issue of single-dimensional input through the addition of an output convolution layer. Furthermore, \cite{Heye2017} employed a recursive decoder prediction method in the encoder-decoder architecture to enhance the accuracy of long-term forecasts. To further show the learning capacity of ConvLSTMs, \cite{Jing2019} implemented a Generative Adversarial Neural Network (GAN) with ConvLSTM units to extrapolate echos to a higher resolution by using adversarial loss. Moreover, U-Net demonstrated high performance in precipitation nowcasting \cite{Agrawal2019, ref2}, achieving results comparable to those of ConvLSTMs.

Attention mechanisms in the spatio-temporal domain are studied in various tasks. For human action recognition, \cite{Song2017} developed temporal and spatial attention mechanisms with fully connected NNs and LSTMs. Similarly, \cite{dual_attn} implemented a dual attention mechanism in time series forecasting. \cite{Liang2018} implemented a multi-level attention mechanism fusing the side-information with multiple geo-sensory time series data to forecast air quality. For traffic prediction, \cite{Yao2019} implemented an attention mechanism on the outputs of LSTM focusing on different time periods from the past. Similarly, \cite{ref1} combined a specific local attention memory mechanism with LSTM to capture spatio-temporal dependencies in precipitation nowcasting. Although these works successfully implement an attention mechanism, the input is bound to be single or multiple time series rather than 3D tensors. As a solution, \cite{Zhang2018} implemented an attention mechanism that took 3D tensors as input, however, the performance increase was limited. Recently, \cite{Abdellaoui2020} conducted a comparison study between an LSTM built on top of a CNN block (Conv+LSTM) and a ConvLSTM with temporally unrolled and temporally concatenated architectures, where the ConvLSTM outperformed the Conv+LSTM configuration.

Finally, to facilitate a comprehensive evaluation and comparison of recently emerging deep learning models, \cite{Rasp2020} introduced a benchmark dataset called ERA5, which is based on predicting global weather patterns in advance. In addition, the dataset involves baseline scores from simple linear regression techniques, deep learning models and purely physical forecasting models.

\subsection{Contributions}
Our main contributions are as follows:
\begin{itemize}
    \item We introduce a novel attention-based recurrent network architecture for spatio-temporal time series called the Weather Model, which incorporates convolution operations on attention layers and utilizes ConvLSTMs as memory units. Our model focuses on the relevant cells in exogenous series to successfully learn spatial correlations and make accurate predictions.

    \item To the best of our knowledge, for the first time in the literature, we construct a specific attention-based ConvLSTM network for spatio-temporal weather forecasting. Our model architecture allows us to input multiple exogenous series and produce refined predictions with high interpretability for each series. In addition, our model is easily applicable to various scenarios, as we can make hourly, daily or any temporal resolution forecasts.

    \item We compare our Weather Model with the state-of-the-art deep learning models in the literature, as well as physical models, in order to illustrate the superiority of our approach. We showcase our results through extensive experiments over the well-known real-life datasets.

    \item We publicly share the implementation of our method for both model design, comparisons and experimental reproducibility \footnote{https://github.com/sftekin/ieee\_weather}.

\end{itemize}

\section{Problem Description}

In this paper, all vectors are real column vectors and are presented in boldface lowercase letters. We represent matrices as bold upper letters and denote the multidimensional tensors by uppercase letters with calligraphic fonts such as $\mathcal{X}, \mathcal{Y}, \mathcal{W}$. The time index is given as subscript and superscript refers to other signals, for example, $\mathbf{X}_{t}^{l}$ shows $l^{\mathrm{th}}$ input signal at the time step $t$ . In addition, we show the entries of a matrix with lowercase letters with subscripts referring to its position, for example, $a_{i,j}$ shows the entry of matrix $\mathbf{A}$ at the $i^{\mathrm{th}}$ row and $j^{\mathrm{th}}$ column.

Our goal is to forecast the weather value of multiple locations using multiple features of historical numeric weather data. The input is the weather features with the grid representation creating a set of spatio-temporal time series. The weather values of a feature $l$ at a time step $t$ can be illustrated as a matrix, $\mathbf{X}_{t}^{l} = \{x_{1,1,t}^l, \dots, x_{i,j,t}^l, \dots, x_{M,N,t}^l\} \in \mathbb{R}^{M\times N}$, where $M$ and $N$ define the spatial dimensions and  $x_{i,j,t}^{l}$ represents the cell value in the grid at the $i^{\mathrm{th}}$ row and $j^{\mathrm{th}}$ column. Following this definition, for a given time window $T$, $\mathcal{X}^{l}=\{\mathbf{X}_{1}^{l}, \dots \mathbf{X}_{T}^{l}\}\in \mathbb{R}^{M\times N\times T}$ represents the spatio-temporal time series for the weather feature $l$. Similarly, for the total number of $L$ features, $\mathcal{X}_{t}=\{\mathbf{X}_{t}^{1}, \dots, \mathbf{X}_{t}^{L}\}\in \mathbb{R}^{M\times N\times L}$ represents all weather features at time step $t$. Altogether, we represent the set of spatio-temporal time series as $\mathcal{X} \in \mathbb{R}^{M\times N \times T \times L}$, capturing all of the grid information of $L$ features until time $T$. Next, we define $\mathbf{Y}_{t}\in \mathbb{R}^{M\times N}$ as our endogenous variable and we represent the target series as $\mathcal{Y}=\{\mathbf{Y}_{1}, \dots, \mathbf{Y}_{T}\}\in \mathbb{R}^{M\times N \times T}$.

Here, our goal is to predict the next sequence of the target series given the previous samples of the input sequences. Mathematically, our objective is to learn the mapping
\begin{align*}
    & \{\mathcal{X}_{1}, \dots, \mathcal{X}_{T_{\mathrm{in}}}\}\xrightarrow[]{\text{g(.)}} \{\mathbf{Y}_{T_{\mathrm{in}}+1}, \dots, \mathbf{Y}_{T_{\mathrm{in}}+T_{\mathrm{out}}}\},
\end{align*}
where $T_{\mathrm{in}}$ and $T_{\mathrm{out}}$ are the input and output sequence lengths, respectively. Note that $T_{\mathrm{out}}$ can be any number of time steps, e.g., we can get one-step ahead prediction, $T_{\mathrm{out}}=1$, or multi-step ahead prediction, $T_{\mathrm{out}}=7$, etc. We suffer the Mean Squared Error (MSE) given as
\begin{align}
    & \mathcal{L}=\frac{1}{T_{\mathrm{out}}MN}\sum_{t=T_{\mathrm{in}}+1}^{T_{\mathrm{in}}+T_{out}}\sum_{i=1}^{M}\sum_{j=1}^{N} (y_{i,j,t}-\hat{y}_{i,j,t})^2,
    \label{eq:loss}
\end{align}
where $\hat{y}_{i,j,t}$ and $y_{i,j,t}$ represent the predicted and ground truth cell values, respectively. Our goal is to minimize the loss \eqref{eq:loss}. Although we use the MSE as the loss measure since it is widely used in the community, any other differentiable loss function can be readily used instead of the MSE in our algorithm as shown in the paper.

\section{The Weather Model}
In this section, we explain our Weather Model architecture, referring to every step and formulation as illustrated in Fig. \ref{fig:all_model_fig}.

\subsection{Capturing Spatio-Temporal Relations with ConvLSTMs}
One of the critical temporal models in deep learning is the RNN \cite{elman}. RNNs store the temporal information of the input sequence within their hidden state vectors, i.e., the memory. The ability to carry information through time in their memory makes RNNs successful in time series forecasting. However, RNNs only have one hidden vector to store the temporal information, which limits their ability to capture long-term dependencies \cite{lstm}. Moreover, RNNs suffer from the well-known vanishing gradient problem \cite{dual_attn}. As a remedy, \cite{lstm} introduced LSTMs, which contain two hidden vectors and an additional memory vector to store information. Thus, LSTMs are more successful in modeling long-term dependencies compared to RNNs. Furthermore, ConvLSTM improves the setup of the LSTMs by integrating convolution operations to the input and the hidden state transitions. In addition, the ConvLSTM architecture can take multiple time series as input to produce outputs for each series simultaneously.

We use ConvLSTMs to capture the spatial correlations present in the weather data, through the use of convolution operations. Thus, we incorporate a specific ConvLSTM architecture into our Weather Model to capture the relation between adjacent grids in the spatio-temporal time series. Each ConvLSTM unit in our Weather Model has inputs $\mathcal{X}_{t}=\{\mathbf{X}_t^{1}, \dots, \mathbf{X}_t^{L_{\mathrm{in}}}\}\in \mathbb{R}^{M\times N\times L_{\mathrm{in}}}$, cell outputs $\mathcal{S}_{t}=\{\mathbf{S}_t^{1}, \dots, \mathbf{S}_t^{L_{\mathrm{hid}}}\}\in \mathbb{R}^{M\times N \times L_{\mathrm{hid}}}$, hidden states $\mathcal{H}_{t}=\{\mathbf{H}_t^{1}, \dots, \mathbf{H}_t^{L_{\mathrm{hid}}}\}\in \mathbb{R}^{M\times N \times L_{\mathrm{hid}}}$ and gates $i_t$, $f_t$, $o_t$, where $L_{\mathrm{in}}$, $L_{\mathrm{out}}$ and $L_{\mathrm{hid}}$ specify the input, output and hidden dimensions, respectively. $\mathcal{X}_t$ and $\mathcal{H}_t$ are the results of 2D convolutions in the input gates shown in Fig. \ref{fig:convlstm}. The update equations of the gates in ConvLSTM are given as
\begin{align*}
    &i_t = \sigma(\mathcal{W}_{xi} \ast \mathcal{X}_t + \mathcal{W}_{hi} \ast \mathcal{H}_{t-1} + \mathcal{W}_{si} \odot \mathcal{S}_{t-1} +  \mathbf{b}_i),\\
    &f_t = \sigma(\mathcal{W}_{xf} \ast \mathcal{X}_t + \mathcal{W}_{hf} \ast \mathcal{H}_{t-1} + \mathcal{W}_{sf} \odot \mathcal{S}_{t-1} +  \mathbf{b}_f),\\
    &\mathcal{S}_t = f_t \odot \mathcal{S}_{t-1} + i_t \odot \mathrm{tanh}(\mathcal{W}_{xs} \ast \mathcal{X}_t + \mathcal{W}_{hs} \ast \mathcal{H}_{t-1} + \mathbf{b}_s),\\
    &o_t = \sigma(\mathcal{W}_{xo} \ast \mathcal{X}_t + \mathcal{W}_{ho} \ast \mathcal{H}_{t-1} + \mathcal{W}_{so} \odot \mathcal{S}_{t} +  \mathbf{b}_o),\\
    &\mathcal{H}_t = o_t \odot \mathrm{tanh}(\mathcal{S}_t),
\end{align*}
where $\ast$ and $\odot$ are the convolution and element-wise multiplication operators, respectively. $\mathcal{W}_{x\sim} \in \mathbb{R}^{I \times J \times L_{\mathrm{in}} \times L_{\mathrm{hid}}}$, $\mathcal{W}_{h\sim} \in \mathbb{R}^{I \times J \times L_{\mathrm{out}} \times L_{\mathrm{hid}}}$, $\mathcal{W}_{s\sim} \in \mathbb{R}^{I \times J \times L_{\mathrm{hid}} \times L_{\mathrm{hid}}}$ and $\mathbf{b}_i, \mathbf{b}_f, \mathbf{b}_o \in \mathbb{R}^{L_{\mathrm{hid}}}$ are the parameters to learn and $I$, $J$ are the height and width of the filters, respectively.

In our Weather Model, we observe the movement patterns in weather features, e.g. temperature, wind and humidity due to the generic atmospheric circulation, through the use of ConvLSTMs. Here, we regard the different states present in the ConvLSTM units as the hidden representation of the aforementioned movement patterns, where we employ the filters of ConvLSTM to capture these patterns. As illustrated in Fig. \ref{fig:model_graph}, in order to extend the learning capacity of our Weather Model, we incorporate multiple ConvLSTM units into our framework in a stacked manner.

\subsection{Interpreting the Learned Information from ConvLSTMs}

In order to interpret the information learned by the ConvLSTM units, we introduce an encoder-decoder structure and incorporate it into our Weather Model. The encoder part of our model processes the input data through the use of ConvLSTMs and generates a fixed-length context vector that captures all the relevant information contained in the input data. Next, the decoder part interprets the knowledge of the context vector and generates the output data, which is the predicted target sequence. Here, we additionally integrate an attention mechanism into the encoder architecture and a context matcher mechanism into the decoder architecture to further boost their performances, which we support through an extensive set of simulations. Next, we thoroughly explain our encoder and decoder architectures.

\subsubsection{Encoder and Attention Mechanism}
We introduce an encoder structure to model the spatio-temporal sequences as shown in Fig. \ref{fig:model_graph}. For $L$ number of spatio-temporal input sequences $\tilde{\mathcal{X}_{t}}=\{\tilde{\mathbf{X}}_t^{1}, \dots, \tilde{\mathbf{X}}_t^{L}\}$ our encoder learns the following mapping
\begin{equation}
    \mathcal{H}_t, \mathcal{S}_{t} = f(\tilde{\mathcal{X}}_{t}, \mathcal{H}_{t-1}, \mathcal{S}_{t-1}),
    \label{eq:rnn2}
\end{equation}
where $\mathcal{H}_t, \mathcal{S}_t \in \mathbb{R}^{M \times N \times L_{\mathrm{hid}}}$ are the hidden states of the encoder at time $t$ and $\tilde{\mathcal{X}}_t$ is the output of the attention mechanism. Here, $\mathcal{S}_{t}$ is omitted from Figs. \ref{fig:model_graph} and \ref{fig:enc_dec} for clarity, in which $\mathcal{H}_{t}$ and $\mathcal{D}_{t}$ represent the hidden states of our encoder and decoder, respectively. The function $f$ in \eqref{eq:rnn2} represents the stacked ConvLSTM units of the encoder. Our encoder recursively performs mapping by using the output states of the previous time steps. We illustrate the unrolled view of the encoder in Fig. \ref{fig:enc_dec} by showing operations at each time step. In each layer, ConvLSTM units encode the input spatio-temporal time series by referring to the hidden state at the previous time step. As given in Fig. \ref{fig:model_graph}, after the input layer, each subsequent layer takes the previous layer's output as input to their ConvLSTM unit. Each layer has unique hidden states $\mathcal{H}_t^k, \mathcal{S}_t^k$, where $k$ represents the layer index. After the input sequence passes through all of the encoder layers, the hidden states $\mathcal{H}_t^k, \mathcal{S}_t^k$ at each encoder layer are passed to the context matcher mechanism, which is the final step of the encoding operation.

To further boost the accuracy of the encoder model, we introduce a specific attention mechanism for this particular spatio-temporal problem. Our attention mechanism helps the encoder to extract relevant information from the spatio-temporal time series at each time step by referring to the hidden state of the previous encoder. As shown in Fig. \ref{fig:model_graph}, for each time step $t$, the attention mechanism takes the input features of the spatio-temporal time series $\mathcal{X}_{t}=\{\mathbf{X}_{t}^{1}, \dots, \mathbf{X}_{t}^{l}, \dots, \mathbf{X}_{t}^{L}\}$ and the encoder's first layer of the previous hidden state $\mathcal{H}_{t-1}^1$ to calculate the attention weights for each feature $l$ given as
\begin{equation}
    \mathbf{E}_t^l = \mathcal{V} \, \mathrm{tanh}(\mathcal{W}_E \ast \mathcal{H}_{t-1}^1 + \mathcal{U} \ast \mathbf{X}_t^{l} + \mathbf{b}_{E}),
    \label{eq:energy}
\end{equation}
where $\mathcal{W}_E \in \mathbb{R}^{I\times J\times L_{\mathrm{hid}}\times L_{\mathrm{att}}}$, $\mathcal{U} \in \mathbb{R}^{I\times J\times L_{\mathrm{in}} \times L_{\mathrm{att}}}$, $\mathcal{V} \in \mathbb{R}^{I\times J\times L_{\mathrm{att}} \times 1}$ and $\mathbf{b}_{E}\in \mathbb{R}^{L_{\mathrm{att}}}$ are the parameters to learn, $L_{\mathrm{att}}$ specifies the attention dimension size and $I$, $J$ are the height and width of the filters, respectively. The output dimension of \eqref{eq:energy} is $M\times N\times 1$ and $\mathbf{E}_{t}^{l}=\{e_{t,1,1}^{l}, \dots, e_{t,i,j}^{l}, \dots e_{t,M,N}^{l}\}$ is called the energy matrix for the feature $l$ at time step $t$. We perform $\mathrm{softmax}$ operation on the energy matrices of each feature to obtain the attention matrix $\mathbf{A}_{t}^{l} = \{a_{t,1,1}^{l}, \dots, a_{t,i,j}^{l}, \dots a_{t,M,N}^{l}\}$, where the $\mathrm{softmax}$ operation at each entry of the attention matrix is given as
\begin{equation*}
    a_{t,i,j}^{l} = \frac{\mathrm{exp}({e}^l_{t,i,j})}{\sum_{l}^{L}\mathrm{exp}({e}^l_{t,i,j})}.
\end{equation*}
Next, we obtain the weighted input spatio-temporal time series by taking the Hadamard product of each feature matrix with the corresponding attention matrix:
\begin{equation*}
    \tilde{\mathbf{X}}_{t} = \{\mathbf{A}_{t}^{1} \odot \mathbf{X}_{t}^{1}, \dots, \mathbf{A}_{t}^{L} \odot \mathbf{X}_{t}^{L}\},    
\end{equation*}
where $\mathbf{A}_{t}^{l}, \mathbf{\tilde{X}}_{t}^{l}\in \mathbb{R}^{M\times N}$. Each cell in $\mathbf{A}_{t}^{l}$ represents the attention value assigned to the respective features of that particular cell. Through the integration of the attention mechanism, our encoder is able to selectively focus on different spatio-temporal time series at each time step. Note that the $\ast$ operation in \eqref{eq:energy} corresponds to 2D convolution. We integrate convolutional operations into the attention mechanism as the fully-connected methods require 2D inputs. In essence, without the convolution operations we would have to flatten the input tensor, hence, lose the information regarding the spatial covariances that we capture through our ConvLSTM units.

\subsubsection{Decoder and Context Matcher Mechanism}
We present a specific decoder architecture to extract the information encoded within the context vector produced by the encoder. As illustrated in Fig. \ref{fig:model_graph}, our decoder consists of stacked ConvLSTM units, same as the encoder. To prevent performance loss given long-length inputs, we additionally incorporate a context matcher mechanism. This mechanism aggregates the hidden states of the encoder layers, given by $\mathcal{H}_1^k + \dots + \mathcal{H}_T^k$. Thus, we can increase the length of the gradient flow up to the first time step. As shown in Fig. \ref{fig:model_graph}, we use symmetry in our architecture and set the number of layers in the encoder and the decoder to be the same. The encoder's hidden state dimensions decrease as the number of layers increases and vice-versa for the decoder. We describe our context matcher mechanism as
\begin{equation*}
    \mathcal{D}_t^{K-k+1} = \sum_{i=t-T_{\mathrm{in}}}^{t} \mathcal{H}_i^k,
\end{equation*}
where $\mathcal{D}_t^{K-k+1}\in \mathbb{R}^{M\times N \times L_{\mathrm{hid}}}$ is the decoder hidden state, $K$ is the total number of layers and $k \in \{1, \dots K\}$. As shown in Fig. \ref{fig:model_graph}, the context matcher sums the hidden states and reverses the order of the states. Here, we reverse the matching of the hidden states of the encoder's layers to increase the performance of the mechanism. Next, the produced state $\mathcal{D}_{t-1}^{1}$ is passed to the decoder's corresponding layers and the input $\mathbf{Y}_t$ passes the first layer of the decoder, as shown in Fig.\ref{fig:model_graph}. The first layer produces the next hidden state $\mathcal{D}_{t+1}^{1}$ and passes it to the next layer. The decoder's final layer produces the input to the output convolutions, $\mathcal{D}_{t+1}^{K}$, as shown in Fig. \ref{fig:model_graph}. The output convolutions produce the prediction $\hat{\mathbf{Y}}_{t+1}$ for the time step $t+1$. Through this approach, we can utilize higher dimensions in the initial layer of the encoder, thereby achieve predictions with any desired number of dimensions. The recursive prediction of the decoder is given in Fig. \ref{fig:enc_dec}. The decoder uses the predictions produced in the previous time step, allowing us to set the forecast horizon to any value.

\subsection{Evaluation Metrics}
During our experiments, we use the benchmark metrics given in \cite{Rasp2020}, for fair comparison. Following their application, we also weigh our evaluation metrics with a predefined latitude weight factor. Since the distance between longitudes decreases as we move toward the poles, the area of each cell in the grid is not equal. The cells that are close to the poles have less area than the cells that are close to the Equator. Thus, we weigh our metrics according to the cell sizes. For the $j^{\mathrm{th}}$ latitude, the weight factor is given as
\begin{equation*}
    l(j) = \frac{cos(lat(j))}{\frac{1}{N}\sum_{j=1}^{N}cos(lat(j))},
\end{equation*}
where $N$ represents the number of latitudes and $lat(j)$ represents the $j^{\mathrm{th}}$ latitude. Our primary metric is the Root Mean Square Error (RMSE) since it reflects our loss function in \eqref{eq:loss} and minimizes the effect of the outliers. We define the latitude-weighted RMSE as
\begin{equation}
    \mathrm{RMSE} = \frac{1}{T_{\mathrm{out}}}\sum_{t=T_{\mathrm{in}}+1}^{T_{\mathrm{in}}+T_{\mathrm{out}}}\sqrt{\frac{1}{MN}\sum_{i=1}^{M}\sum_{j=1}^{N}l(j)(\hat{y}_{i,j,t} - y_{i,j,t})^2},
    \label{eq:rmse}
\end{equation}
where $T_{\mathrm{out}}$ represents the output sequence length. We use latitude-weighted Mean Absolute Error (MAE) as our second evaluation metric, where we take the absolute value of the error instead of taking the square in \eqref{eq:loss} and perform the weight operation in the same way shown in \eqref{eq:rmse}. We choose the latitude-weighted anomaly correlation coefficient (ACC) as our final metric, defined as
\begin{equation}
    \mathrm{ACC} = \frac{\sum_{i,j,t}l(j)y^{'}_{i,j,t}\hat{y}^{'}_{i,j,t}}{\sqrt{\sum_{i,j,t}l(j)(y^{'}_{i,j,t})^2\sum_{i,j,t}l(j)(\hat{y}^{'}_{i,j,t}})^2},
\end{equation}
where $'$ denotes the difference to the climatology, $y^{'}_{i,j,t} = y_{i,j,t} - \bar{y}_{i,j}$, where $\bar{y}_{i,j}$ represents the climatology and is defined as
\begin{equation*}
    \bar{y}_{i,j} = \frac{1}{T_{\mathrm{out}}}\sum_{t}y_{i,j,t}.
\end{equation*}
The ACC metric takes values between $+1$ and $-1$, where positive values show that the forecast anomalies have represented the observed anomalies successfully. 

\subsection{The Prediction Methods}\label{subsec:pred_methods}
In our simulations, we forecast multiple time steps ahead of the target endogenous series. For example, if the temporal resolution is hourly and we want to forecast for the three-day interval, we predict for the following $72$ steps. We obtain these predictions in three methods. In our first method, sequential prediction, we set the output length $T_{\mathrm{out}}$ to $72$ in our model, then the decoder produces the output for each time step $t > T_{\mathrm{in}}$. In our second method, iterative prediction, we set the output length $T_{\mathrm{out}}$ to $6$, then use the predictions as inputs for the next 6 hours and repeat this process for 12 iterations. However, if the model uses four exogenous series as input, we additionally need to predict the four series at each iteration because we assume that we also do not know the future values of the given exogenous series. In our third method, direct prediction, we directly perform forecasts for the $n^{\mathrm{th}}$ time step. For example, if we want to predict only the $24^{\mathrm{th}}$ step of the endogenous series, we set the temporal resolution to 24 hours and predict for one step, i.e., $T_{\mathrm{out}}=1$. In our experiments, we use all three of our prediction methods and compare their performance on two different datasets. The following section focuses on the datasets that we employ and our experimental results.

\section{Experiments}
We perform two sets of experiments using two different ERA5 reanalysis datasets. In the first setup, we asses our model on a high spatial resolution dataset covering Turkey by comparing it with the state-of-the-art deep learning models. In the second setup, we compare our model performance with the state-of-the-art deep learning models and physical models on a benchmark dataset called WeatherBench \cite{Rasp2020}.
\subsection{Data Description}
\subsubsection{The Weather Dataset}
Our first dataset is the ERA5 hourly data on pressure levels \cite{Hersbach2020}, where the temporal resolution of the data is $3$ hours, the spatial resolution is $30$ kilometers and the data belongs to the $100$ hPa atm pressure level. The grid dimensions of the data are $(61\times121)$. Our data captures the dates between $2000$ and $2001$, where the spatial range is $30^\circ$ - $45^\circ$ in latitude and $20^\circ$ - $50^\circ$ longitude covering the Mediterranean and Black Sea Coasts of Turkey. In Table \ref{table:features}, we illustrate the weather features available for use in the dataset. Among these features, we select the temperature as our endogenous series and the remaining features as our exogenous series. Thus, our task is to find the next temperature values using the previous values of the weather features. We split the dataset into the train, validation and test sets, with $0.8$, $0.1$ and $0.1$ split ratios, respectively.

\begin{table*}[t]
    \caption{The description and corresponding units of the features present in the ERA5 dataset.}
    \label{table:features}
    \vspace*{-2mm}
    \begin{center}
        \begin{tabular}{cccc}
            \toprule
            \textbf{Long Name} & \textbf{Short Name} & \textbf{Description}  & \textbf{Unit} \\
            \midrule
            geopotential & z & The gravitational potential energy of a unit mass & $\mathrm{m^2s^{-2}}$ \\
            \hline
            potential vorticity & pv & The measure of the capacity for air to rotate in the atmosphere & $\mathrm{Km^2kg^{-1}s^{-1}}$ \\
            \hline
            relative humidity & r & \makecell{The water vapor pressure as a percentage of the value at which \\ the air becomes saturated} & $\mathrm{\%}$ \\
            \hline
            specific humidity & s & The mass of water vapour per kilogram of moist air & $\mathrm{kg\:kg^{-1}}$ \\
            \hline
            temperature & t & Temperature & $\mathrm{K}$ \\
            \hline
            u-wind & u & The eastward component of the wind & $\mathrm{ms^{-1}}$ \\
            \hline
            v-wind & v & The northward component of the wind & $\mathrm{ms^{-1}}$ \\
            \hline
            vertical velocity & w & The speed of air motion in the upward or downward direction & $\mathrm{Pa}\: s^{-1}$ \\
            \bottomrule
        \end{tabular}
    \end{center}
\end{table*}

\subsubsection{The Benchmark Dataset}
Our second dataset is WeatherBench, which is introduced and explained in \cite{Rasp2020}. The grid size of the data is $(32\times64)$ and the temporal resolution is hourly. The $850$ hPa temperature feature is selected as our endogenous series, whereas the remaining $18$ features are selected as our exogenous features. We obtain direct and iterative forecasts for $3$-$5$ days of lead time. For the training and validation, we use the data between $2015$ and the end of $2016$, and separate the data between $2017$ and $2018$ as the test dataset.

\subsection{Baseline Models}
We select five deep learning models to compare their performances with our Weather Model.
\begin{itemize}
    \item ConvLSTM: We follow the forecasting network in \cite{convlstm}, which involves an encoder-decoder structure. We select this model as our ablation experiment, as we aim to achieve superior performance using our Weather Model that has attention, context matcher and output convolutions. The encoder consists of three layers of ConvLSTM units. $1, 16, 32$ and $32, 16, 1$ are the number of hidden dimensions in the encoder and the decoder layers, respectively.
    
    \item U-net: We implement the given model architecture due to its success in the recent weather forecasting literature \cite{Ronneberger2015}. The model takes a sequence of spatio-temporal temperature time series as the input and predicts their output sequences. The architecture consists of $4$ downsampling and $4$ upsampling layers.

    \item TrajGRU: TrajGRUs have shown slightly better performance than ConvLSTMs in the recent literature \cite{trajgru}. Thus, we added this model to compare its performance with ConvLSTM and our Weather Model. The encoder-decoder structure is the same as the baseline ConvLSTM and we set the number of links to $1$.

    \item LSTM: We integrate LSTMs into our experimetnal setup in order to compare the performances of the ConvLSTMs with vanilla LSTMs. We flatten the grid at each time step to create $(T, MN)$ input matrices to feed the LSTMs. We set the hidden dimension to $256$. After the LSTM produces $(T, 256)$ output matrices, we pass them to a FC-NN with $(256, MN)$ matrix size. After reshaping the outputs to the grid size, we produce the outputs for the given spatio-temporal time series.
    
    \item Simple Moving Average (SMA): We implement a simple but effective statistical model called the Simple Moving Average (SMA). In this model, we predict the value of the series at time $t$ by taking the weighted average of the spatio-temporal input series values with window length $T$  \cite{time_series_forecasting}. The averaging operation can be performed with equal weights or we can give a higher value to the recent values of the series, such as $t-1,\dots,t-7$. To further boost the performance of the model, we implement an FC-NN to find the optimal weights for each time series. At each epoch, we initialize the weights with a uniform random variable where the range is $[0, 1]$ and optimize them using the FC-NN.
\end{itemize}

\subsection{Hyper Parameter Selection}
To select the best-performing parameters for each model in order to have a fair comparison, we split the datasets into train, validation and test periods. We train our model on the train set and validate it on the validation set to find the best collection of parameters. We then train the models on the train and validation sets combined using the best collection of parameters. Lastly, we test our model on the test data to determine its performance for comparison.

We use the Adam optimizer with $0$ weight decay in each experiment and select the best-performing learning rate from $\{0.01, 0.001, 0.0005, 0.00001\}$. We set the layer size equal for each baseline model and choose the best performing hidden dimension size from \{$16, 32, 64\}$.

\begin{figure*}[t!]
    \centering
    \begin{subfigure}{0.25\textwidth}
        \includegraphics[width=\textwidth]{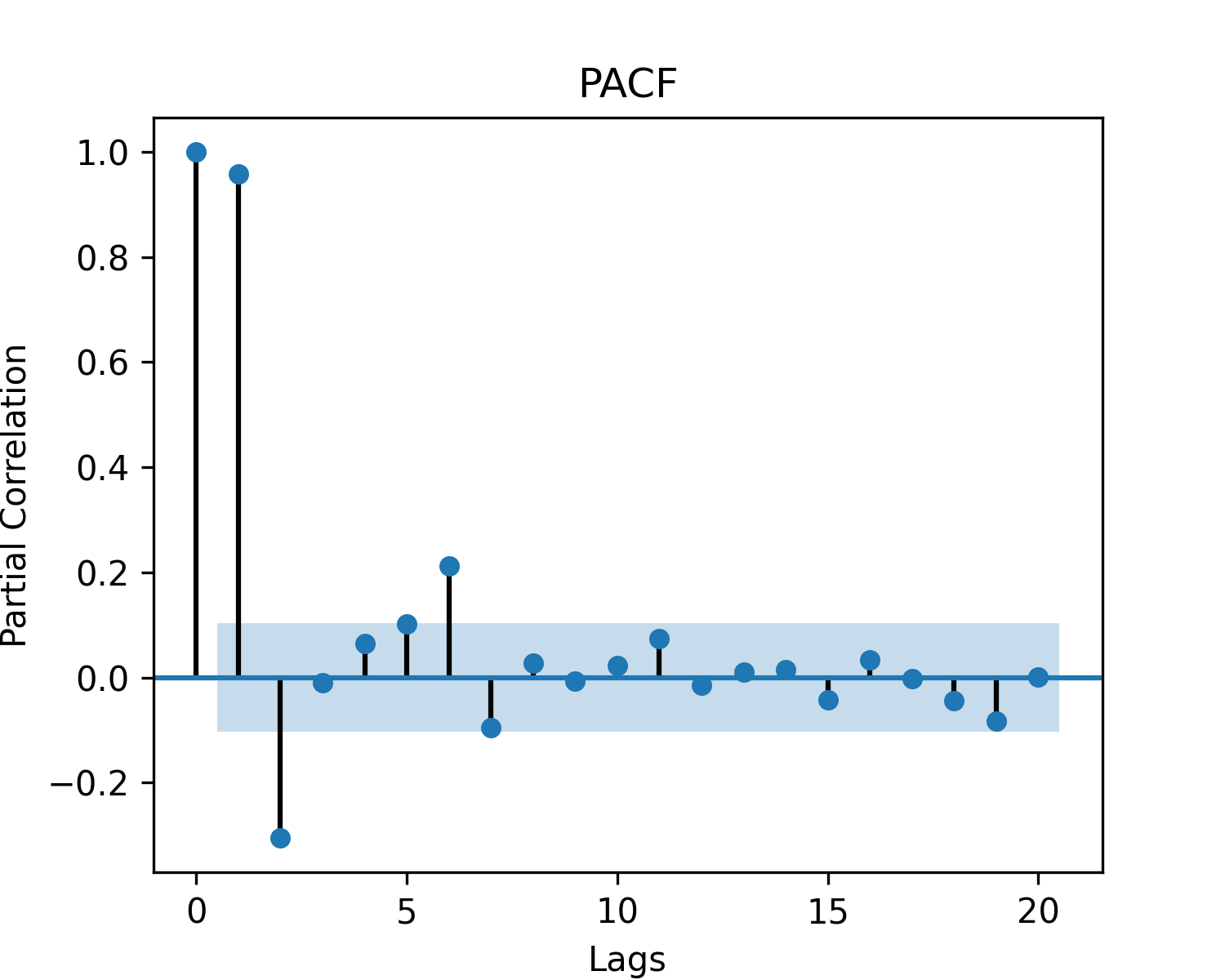}
    \end{subfigure}
    \begin{subfigure}{0.25\textwidth}
        \includegraphics[width=\textwidth]{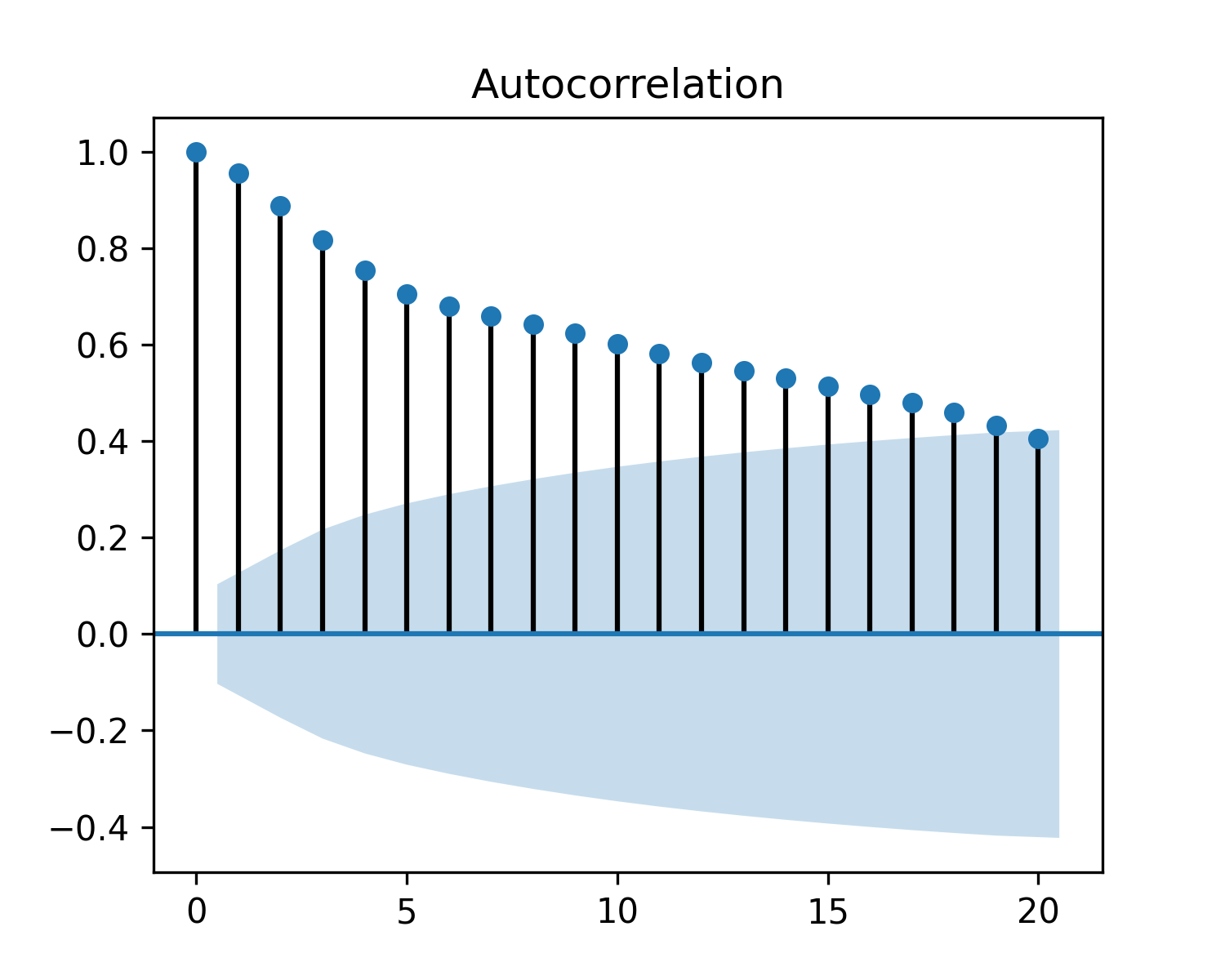}
    \end{subfigure}
    \caption{Partial autocorrelation and autocorrelation functions of temperature value for a randomly selected cell. The blue region shows the $95\%$ confidence level and each lag represents three hours difference.}
    \label{fig:acf_pacf}
\end{figure*}

We performed autocorrelation (ACF) and partial autocorrelation analysis (PACF) on temperature values of a randomly selected cell as shown in Fig. \ref{fig:acf_pacf}. We observed a high correlation until lag $20$ with the ACF. In addition, with the PACF we can find the direct relationship between the observation and its lagged values by removing correlations due to the terms at shorter lags \cite{metcalfe2009introductory}. As illustrated in Fig. \ref{fig:acf_pacf}, we observe high correlation in the first $2$ lags and meaningful correlations until the $10$th lag. Therefore, we selected $10$ time steps of input data with $3$ hours of frequency and predict $10$ time steps ahead.

In addition, input series have different units and scales, hence, requiring normalization. However, the dataset is massive and we need high RAM capacity to normalize a dataset of such size. Thus, we instead apply min-max normalization to the batches in an adaptive manner.

\begin{figure*}[t!]
    \centering
    \begin{subfigure}{0.6\textwidth}
        \centering
        \includegraphics[width=\textwidth]{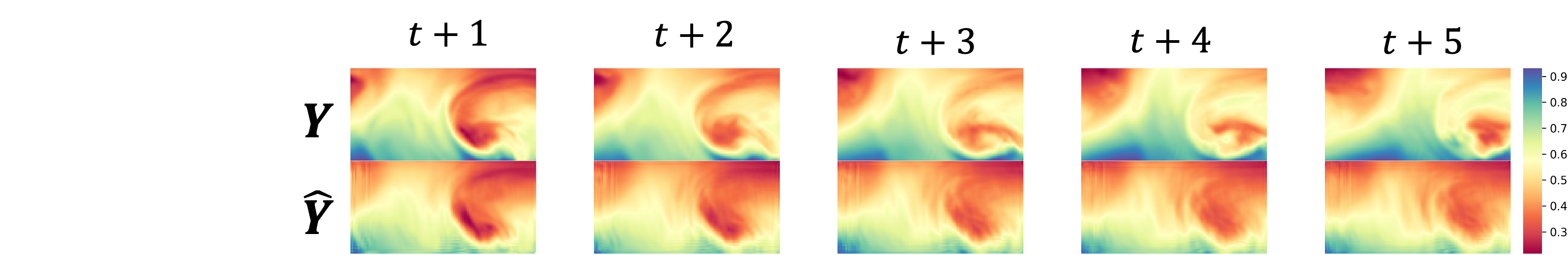}
        \caption{Predicted and ground truth sequence}
        \label{fig:y}
    \end{subfigure}
    \begin{subfigure}{0.6\textwidth}
        \centering
        \includegraphics[width=\textwidth]{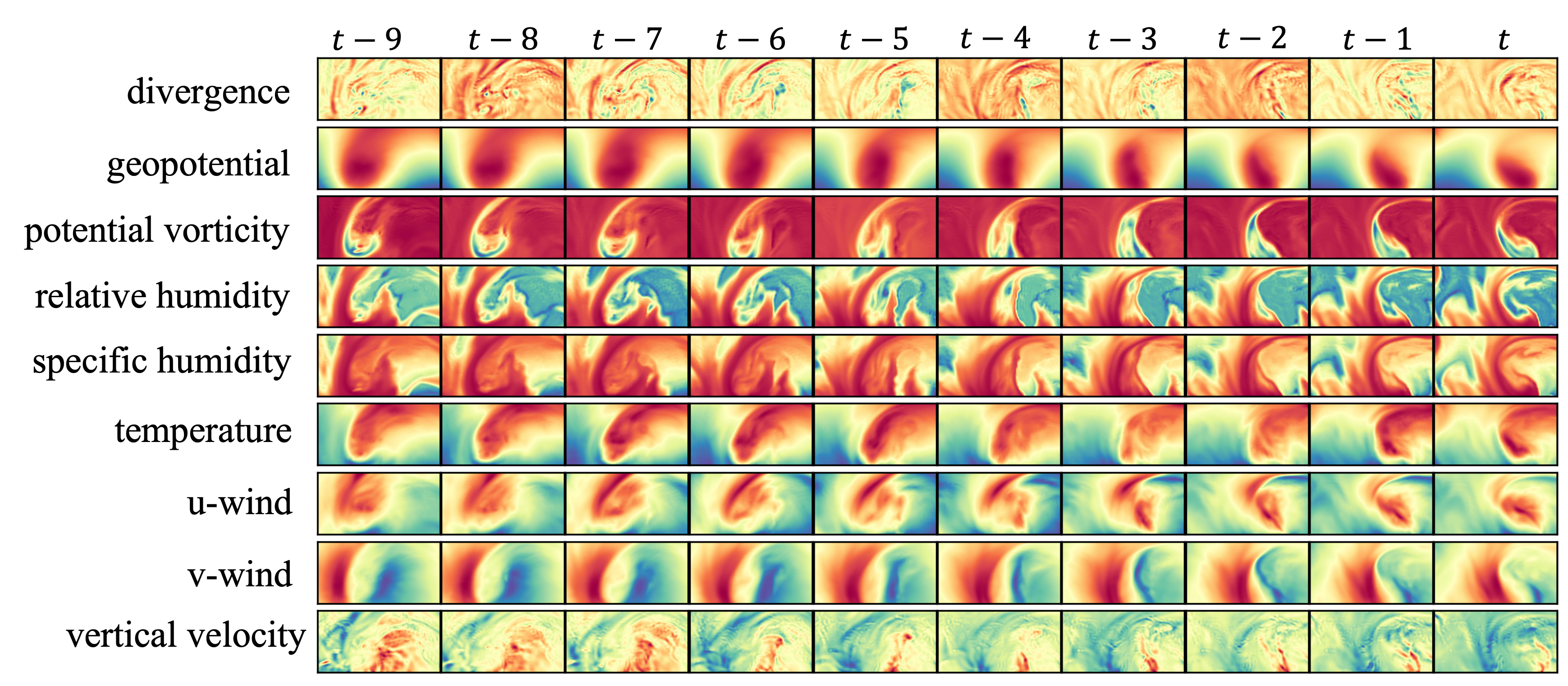}
        \caption{Input sequences}
        \label{fig:x}
    \end{subfigure}
    \begin{subfigure}{0.6\textwidth}
        \centering
        \includegraphics[width=\textwidth]{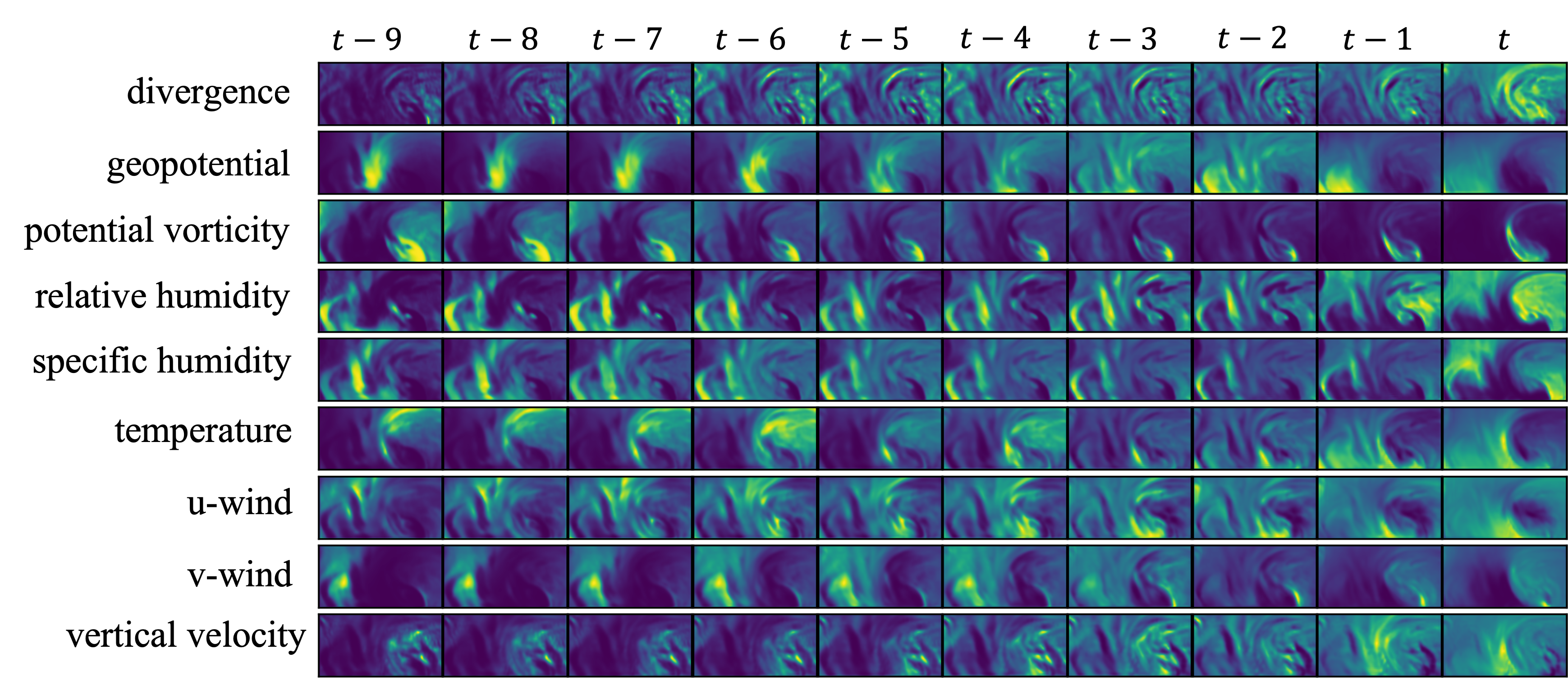}
        \caption{Attention weights}
        \label{fig:alpha}
    \end{subfigure}
	\caption{We show the output sequence (a) of our model along with the input sequences (b) and the attention weights (c) at each time step. At each step, the model focuses on those cells in each feature it finds significant for advancing to the next hidden state. This determination is guided by the input, which corresponds to the temperature value from the previous step. We observe that the model pays attention to the moving parts in each feature, e.g., potential vorticity and geopotential. When we compare the predicted and ground truth series in (a), we observe that the model is successful in the first step. After the first step, the model can forecast only the stationary parts, such as the regions that stay cold or hot. }
	\label{fig:xy_alpha}
\end{figure*}

\subsection{Performance Analysis and Results}
In the first setup of experiments, we evaluate the performance of our model on the high-resolution dataset by comparing it with the baseline deep learning models \cite{convlstm, trajgru, Ronneberger2015}. All the models produce predictions using the sequential method introduced in Section \ref{subsec:pred_methods}, where the output sequence length is set to $5$. We illustrate both numerical and visual results to better assess the model performances. 

The numerical results of the experiment are illustrated in Table \ref{table:results1}. We performed a two-tailed t-test with a p-value of $0.05$ comparing ConvLSTM results with our Weather Model, where the Weather Model is shown to perform significantly better in all metrics. Furthermore, the Weather Model outperforms all of the baseline deep learning models, including TrajGRU and U-Net. When we compare each baseline method, we observe that TrajGRU shows similar performance to ConvLSTM. Furthermore, the models that involve convolution operations, e.g., U-Net, ConvLSTM and TrajGRU, outperform fully connected models, e.g., LSTM and SMA. Thus, our experiments demonstrate the necessity of using convolution operations for accurate modeling of weather movements.

\begin{table}[t]
    \caption{We illustrate the performance of the models on the high-resolution dataset. The best results for each metric are given in bold.}
    \label{table:results1}
    \vspace*{-2mm}
    \begin{center}
        \begin{tabular}{ccccc}
            \toprule
            \textbf{Model Name} & \textbf{RMSE} & \textbf{MAE}  & \textbf{MAPE} \\
            \midrule
            Weather Model & \textbf{1.22115} & \textbf{0.88872} & \textbf{0.00394} \\
            U-net & 1.33032 & 0.97636 & 0.00434 \\
            Convlstm & 1.59211 & 1.18715 & 0.00527 \\
            TrajGRU & 1.60205 & 1.21569 & 0.00540 \\
            SMA & 1.90273 & 1.49958 & 0.00666 \\
            LSTM & 1.99607 & 1.62078 & 0.0072 \\
            \bottomrule
        \end{tabular}
    \end{center}
\end{table}

\begin{table}[t]
    \caption{We illustrate the performance of the models on the WeatherBench dataset. The IFS models are the physical models and their scores, along with the CNN model, are taken from \cite{Rasp2020} for a fair comparison. The best results for each metric are given in bold.}
    \label{table:results2}
    \vspace*{-2mm}
    \begin{center}
        \begin{tabular}{ccccc}
            \toprule
            \textbf{Model Name} & \textbf{RMSE} & \textbf{MAE}  & \textbf{ACC} \\
            \midrule
            Weather Model & \textbf{1.22115} & \textbf{0.88872} & \textbf{0.00394} \\
            ConvLSTM (sequential) & 4.43542 & 3.46913 & 0.63598 \\
            TrajGRU (sequential) & 3.5988 & 2.78978	& 0.75716 \\
            U-net (sequential) & \textbf{3.05466} & \textbf{2.07594} & \textbf{0.82028} \\
            WM (sequential) & 3.97452 & 3.25712 & 0.6909 \\
            \hline
            ConvLSTM (iterative) & 3.16472 & 2.22504 & 0.80173 \\
            TrajGRU (iterative) & 3.89876 & 3.1951 & 0.73826 \\
            U-net (iterative) & 4.13038	& 3.26531 & 0.68035 \\
            CNN (iterative) & 4.48 & 3.49 & 0.72 \\
            WM (iterative) & \textbf{2.91701} & \textbf{2.08413} & \textbf{0.83786} \\
            \hline
            WM (direct) & 3.07334 & 2.2027 & 0.82357 \\
            CNN (direct)  & \textbf{2.87} & \textbf{2.02} & \textbf{0.85} \\
            \hline
            IFS T42 & 3.09 & 1.99 & 0.86 \\
            IFS T63 & 1.85 & 1.30 & 0.94 \\
            Operational IFS & \textbf{1.36} & \textbf{0.93} & \textbf{0.97} \\
            \bottomrule
        \end{tabular}
    \end{center}
\end{table}

\begin{figure*}[!ht]
    \centering
    \begin{subfigure}{0.4\textwidth}
        \includegraphics[width=\textwidth]{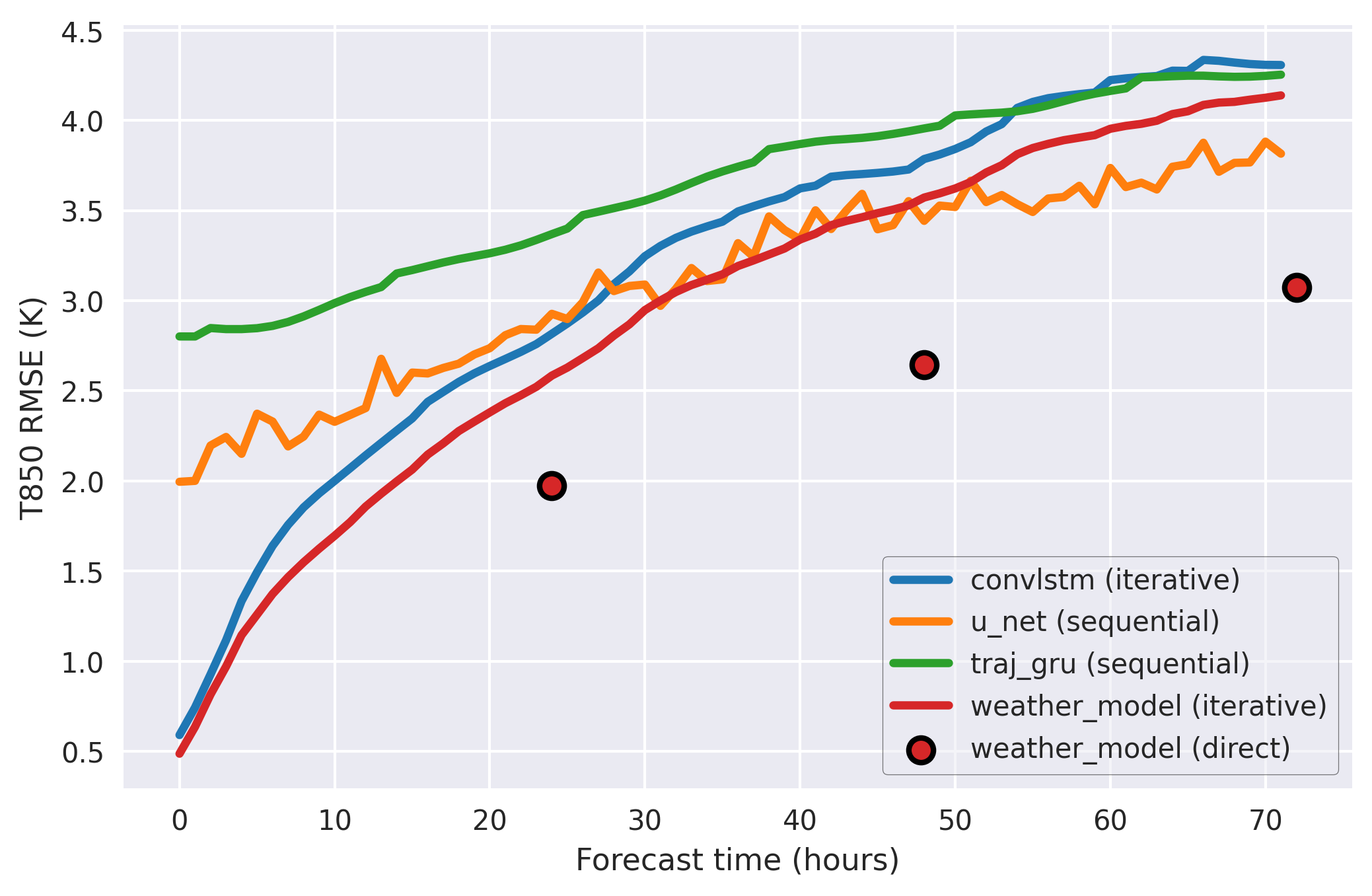}
    \end{subfigure}
    \begin{subfigure}{0.4\textwidth}
        \includegraphics[width=\textwidth]{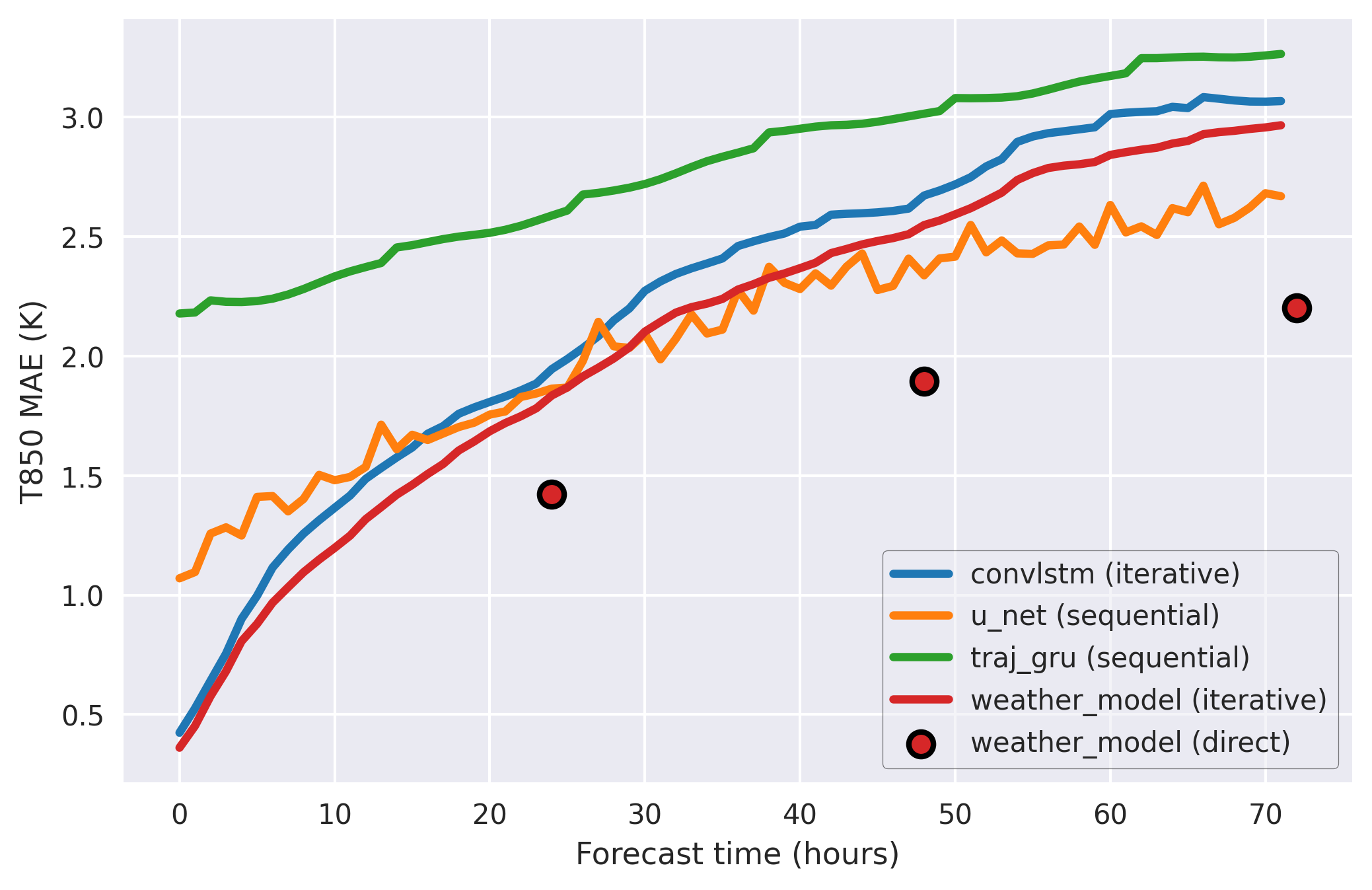}
    \end{subfigure}
    \begin{subfigure}{0.4\textwidth}
        \includegraphics[width=\textwidth]{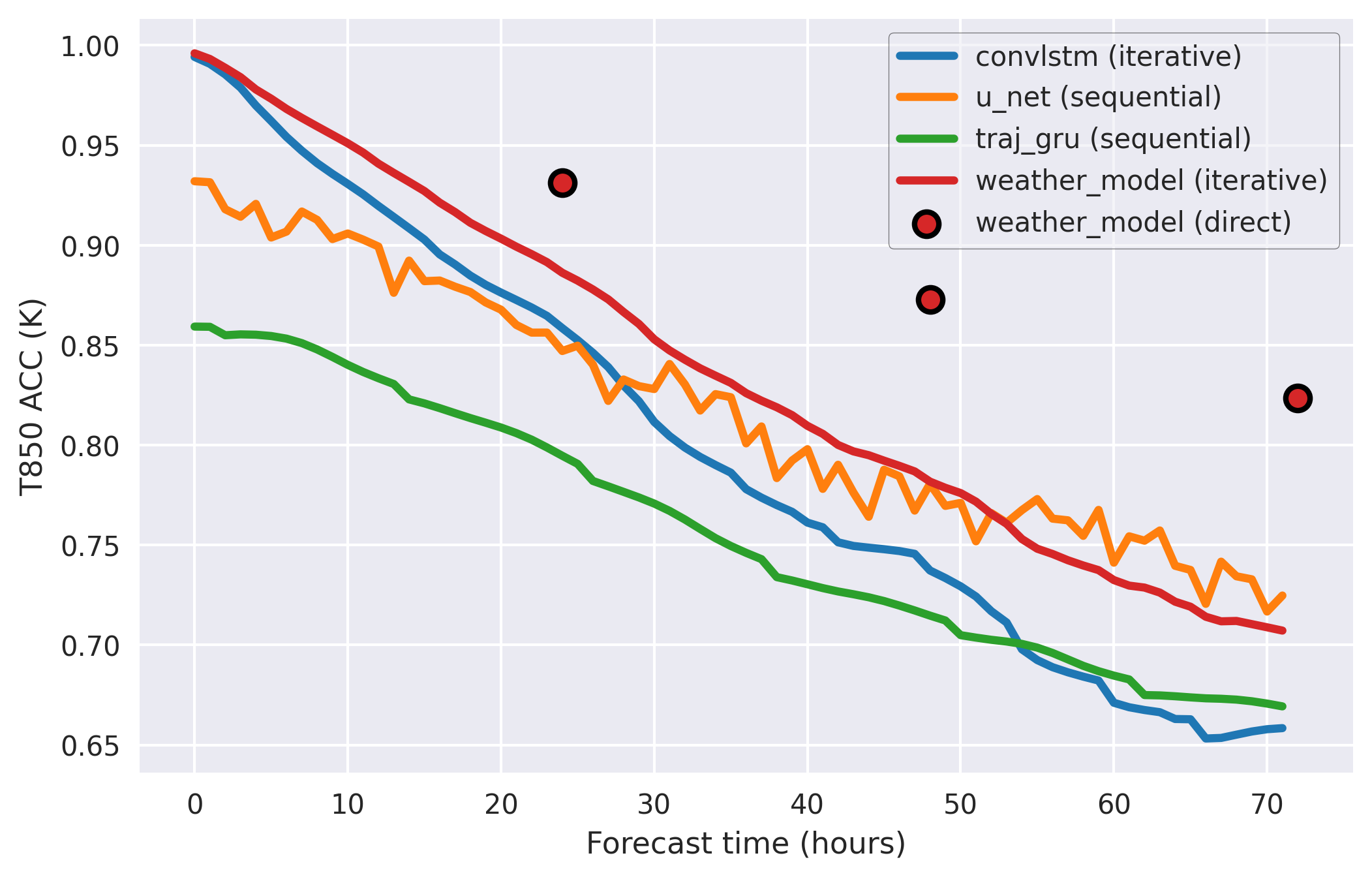}
    \end{subfigure}
    \caption{Evaluation metrics comparison of baseline models for 72 hours of forecast. We show the best-performing method for each model. We also show the results of the direct Weather Model for the given 72 hours of forecast.}
    \label{fig:weighted_metrics}
\end{figure*}

To visually analyze the spatial and temporal performances, we illustrate the predicted and ground truth sequences in Fig. \ref{fig:y}. Observing the changes between the prediction steps, we observe that our Weather Model successfully captures temperature fluctuations up to three steps. However, the performance of the Weather Model deteriorates as the prediction interval length increases. With the help of attention weights shown in Fig. \ref{fig:alpha} and the input series shown in Fig. \ref{fig:x}, we show that our model attends to the movements in the input features and produces the next hidden state by referring to the temperature at the previous step, demonstrating the success of the attention mechanism.

In the second setup of experiments, we evaluate the performance of our Weather Model on the WeatherBench dataset by comparing its performance with the physical and baseline models. Additionally, we compare our results with the CNN model given in WeatherBench. We obtain two different results for each baseline model, one for the sequential and one for the iterative method described in Section \ref{subsec:pred_methods}. In both methods, our goal is to forecast up to $72$ hours. Thus, we set the output sequence length to $72$ and $6$ in sequential and iterative methods, respectively. We show the results in Table \ref{table:results2}. Despite the U-net model obtaining the best scores in the sequential method, our Weather Model trained iteratively obtains the best scores among all deep learning models. We show the hourly performance of the models in Fig. \ref{fig:weighted_metrics}, where the error growth is seen to be logarithmic in RMSE, MAE and linear in ACC metrics. On the other hand, the error growth of the CNN model in \cite{Rasp2020} is exponential. Hence, our Weather Model solves the exponential error growth problem present in \cite{Rasp2020}.

We also perform direct forecasts with our Weather Model and compare its performance with the direct CNN model. We set the temporal resolution to $24$ hours and the output sequence length to $3$. The CNN model performs better than the Weather Model in the direct method. To further increase the performance of the direct Weather Model, we set the temporal resolution to $72$ hours and predict a single step at a time. However, this method performs worse than the previous method. After we observed the input sequence in both direct setups, we came to the conclusion that the spatial movements are lost within significant temporal gaps. Consequently, we conclude that the performance of our Weather Model is contingent upon the spatial movements present in the data.

On the other hand, even though our Weather Model performs better than IFS T42 in the RMSE metric, the physical models significantly outperform all given deep learning models. The reason is that the physical models can simulate the atmosphere better for long forecasts. On the contrary, our Weather Model demonstrates high performance in shorter periods.

\section{Conclusion}
We studied the problem of spatio-temporal weather forecasting and introduced a novel deep learning architecture. We introduced a complete model architecture built on stacked ConvLSTM units with attention and context matcher mechanisms to extend the classical encoder-decoder method. We evaluated our model on high-resolution and benchmark datasets and the results show that our model achieves statistically significant performance gains over the state-of-the-art baseline models. We also showed that the error growth of our model on the long forecasts is logarithmic while it is exponential in the benchmark approach. In addition, we also compared our model with high-resolution physical models and the physical models show superior performance due to their accurate long forecasts, with trade-off of computation time and resources.

\bibliographystyle{elsarticle-num}
\bibliography{main.bib}

\end{document}